\documentclass[sigconf]{acmart}
\AtBeginDocument{%
  }

\copyrightyear{2025}
\acmYear{2025}
\setcopyright{acmlicensed}
\acmConference[MM '25] {Proceedings of the 33rd ACM International Conference on Multimedia}{October 27--31, 2025}{Dublin, Ireland.}
\acmBooktitle{Proceedings of the 33rd ACM International Conference on Multimedia (MM '25), October 27--31, 2025, Dublin, Ireland}
\acmISBN{979-8-4007-2035-2/2025/10}
\acmDOI{10.1145/3746027.3754709}
\usepackage{balance}
\usepackage{graphicx}
\usepackage{epstopdf}
\usepackage{booktabs}
\usepackage{makecell}
\usepackage{color}
\usepackage{caption}
\usepackage{subfigure}
\usepackage{array}

\begin{document}

\title{Chatting about Conditional Trajectory Prediction}

\author{Yuxiang Zhao}
\orcid{0009-0005-3489-705X}
\affiliation{
  \institution{Sun Yat-sen University}
  \city{Shenzhen}
  \country{China}
}
\affiliation{
  \institution{Alibaba Group}
  \city{Beijing}
  \country{China}
}
\email{zhaoyao.zyx@alibaba-inc.com}

\author{Wei Huang}
\authornote{Corresponding Author.}
\orcid{0000-0003-4377-8237}
\affiliation{%
  \institution{Sun Yat-sen University}
  \city{Shenzhen}
  \country{China}
}
\email{huangwei5@mail.sysu.edu.cn}

\author{Haipeng Zeng}
\orcid{0000-0002-0339-0361}
\affiliation{%
  \institution{Sun Yat-sen University}
  \city{Shenzhen}
  \country{China}
}
\email{zenghp5@mail.sysu.edu.cn}

\author{Huan Zhao}
\orcid{0009-0003-2648-0137}
\affiliation{%
 \institution{Sun Yat-sen University}
 \city{Shenzhen}
 \country{China}
}
\email{zhaoh77@mail2.sysu.edu.cn}

\author{Yujie Song}
\orcid{0009-0002-7367-6645}
\affiliation{%
  \institution{Sun Yat-sen University}
  \city{Shenzhen}
  \country{China}
}
\email{songyj28@mail2.sysu.edu.cn}

\renewcommand{\shortauthors}{Yuxiang Zhao, Wei Huang, Haipeng Zeng, Huan Zhao, and Yujie Song}

\begin{abstract}
Human behavior has the nature of mutual dependencies, which requires human-robot interactive systems to predict surrounding agents' trajectories by modeling complex social interactions, avoiding collisions and executing safe path planning. While there exist many trajectory prediction methods, most of them do not incorporate the own motion of the ego agent and only model interactions based on static information. We are inspired by the humans' theory of mind during trajectory selection and propose a \underline{\textbf{C}}ross time domain \underline{\textbf{i}}ntention-interactive method for conditional \underline{\textbf{T}}rajectory prediction(CiT). Our proposed CiT conducts joint analysis of behavior intentions over time, and achieves information complementarity and integration across different time domains. The intention in its own time domain can be corrected by the social interaction information from the other time domain to obtain a more precise intention representation. In addition, CiT is designed to closely integrate with robotic motion planning and control modules, capable of generating a set of optional trajectory prediction results for all surrounding agents based on potential motions of the ego agent. Extensive experiments demonstrate that the proposed CiT significantly outperforms the existing methods, achieving state-of-the-art performance in the benchmarks.
\end{abstract}

\maketitle

\section{Introduction}
Trajectory prediction of surrounding agents is a crucial component for ensuring safety in autonomous driving systems, as it enables avoiding collisions with human-driven cars. Moreover, predicting trajectories of surrounding agents has extensive applications in human-robot interaction, social robots, drones, and other domains\cite{wang2023wsip, mao2023leapfrog, chen2022scept, bae2024singulartrajectory}. Humans can navigate through various social scenarios because they have an intrinsic theory of mind, which is the capacity to reason about other human's actions based on their mental states. Imbuing autonomous systems with such capability could enable more informed decision making and motion planning\cite{song2024robot, zhang2023trajpac, xie2021congestion, gu2022stochastic}. However, predicting trajectory of agents in real world is challenging since an agent’s trajectory is not determined by itself but involves complex social interactions with surrounding agents. Therefore, previous works\cite{gupta2018social, deo2018convolutional, zhou2023query, yan2023int2, guo2022end} have proposed a series of methods to model such interactions.

Despite the remarkable progress have been made, these methods still face three critical problems. First, far less attention is paid to the ego agent's own motion, which hinders direct application of these models to real-world robotic systems. Being able to predict surrounding agents' corresponding reactions based on different potential motions of the ego agent is a very crucial capability for downstream tasks such as decision making, motion planning in robotic systems. For example, when the ego agent is faced with multiple trajectories to choose from, it can generate the predicted future trajectories of the surrounding target agents for each candidate trajectory respectively. Then, the trajectory that optimizes the overall system time/efficiency is selected as the final execution trajectory. Second, they do not conduct dynamic modeling of social interactions over time. During the movement of the agent, its intention dynamically changes as it interacts with surrounding agents. Therefore, trajectory prediction models should jointly analyze intentions over time to achieve dynamic modeling of social interactions. Third, different surrounding agents have varying degrees of influence on the target agent whose trajectory we want to predict. Many convolution and social pooling methods extract features of surrounding agents and directly concatenate them without letting the network learn the degree of influence in a prioritized manner.
\begin{figure}[tbp]
  \centering
   \includegraphics[width=1.0 \linewidth]{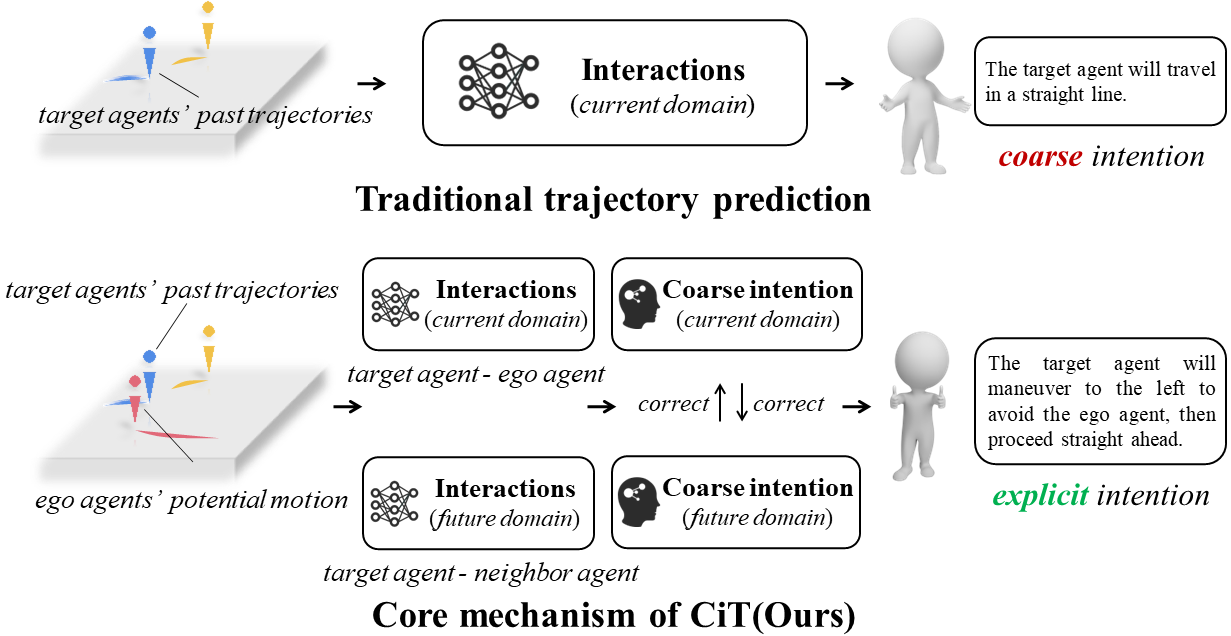}
   \caption{CiT incorporates potential motion of the ego agent to model social interactions, capturing the evolving intentions of the target agent over time. Furthermore, by jointly analyzing the intentions, it achieves feature refinement and information complementarity, leading to clearer and more accurate intention information for trajectory prediction.}
   \label{Fig.1}
\end{figure}
In order to model complex social interactive behaviors more delicately and tightly integrate with robotic system downstream planning and control tasks, we propose the CiT to produce behavior trajectory prediction of all surrounding agents based on ego-agent motion plans. The core of the proposed CiT is to mutually complement and refine intention representations over time through semantic supplementation and feature correction. Figure \ref{Fig.1} illustrates the main working mechanism of the proposed model.

CiT contains four key designs: First, we introduce the future trajectory of the ego agent. Note that CiT does not require the exact future trajectory, which is actual undetermined during prediction. CiT only conditions a rough trajectory which can be easily obtained by trajectory generator. This allows our proposed model to generate optional predictions based on candidate trajectories proposed by downstream planning and control modules. Second, in the intention graph construction module, by analyzing the past trajectories of the target agent and neighbor agents, we can infer the current intention of the target agent. To preserve spatial information, we map this intention onto a social tensor according to the target agent's location and refer to it as the "Intention Graph in the Current Time Domain." Furthermore, by incorporating the future trajectory of the ego agent, we can model the potential social interactions between the target agent and the ego agent in the future and predict the future intention of the target agent. Similarly, we map this future intention onto a social tensor based on its location and refer to it as the "Intention Graph in the Future Time Domain." Third, since the intention information from both time domains during the construction of the intention graph is partial and coarse, in the interaction cross domain module, intention information from different time domains interacts with each other. The intention in one time domain proposes a Query to the other time domain and corrects its own intention through the Key and Value from the other time domain. Through joint analysis of intention information over time, features across different agents, spaces, and time domains are fully extracted and fused to obtain a more precise intention representation. Fourth, in the intention influence evaluation module, the network estimates the degree of influence of different intentions on the future trajectory of the target agent, further refining the interaction process. The main contributions are concluded as follows:
\begin{itemize}
\item We propose CiT, which comprises four novel designs, including 1) future motion incorporation, captures the interactive aspect in human-robot interaction, 2) intention graphs construction, constructs two types of intention graphs, 3) interaction cross domain, achieves information complementarity between the two intentions, integrating information across different time domains and agents, 4) intention influence evaluation, enables the network to consider the degree of influence from different agents in a prioritized manner.
\item In robotic systems, multiple candidate trajectories can be generated to evaluate their corresponding performance in the prediction module, the CiT will provide a highly valuable interface for integrating this trajectory prediction model into robotic system.
\item We conduct experiments on two real-world datasets to evaluate our method. Experimental results show that CiT achieves state-of-the-art performance.
\end{itemize}

\section{Related Work}
\subsection{Trajectory Prediction}
Given an observed past trajectory, trajectory prediction aims to forecast each agent's future trajectory over a period of time. Many scene context-aware methods\cite{liu2024laformer, chen2022scept, wagner2023road} attempt to incorporate bird's eye view road images or lane information as inputs to prune unlikely or low probability trajectories. VectorNet\cite{gao2020vectornet} encodes both image information and each agent as vectors and uses a novel graph convolution network. RedMotion\cite{wagner2023road} proposes a transformer-based model, which learns environment representations through two types of redundancy reduction. LAformer\cite{liu2024laformer}uses an attention-based temporally dense lane-aware estimation module to assess the alignment probability between motion and HD map scene information, enhancing environmental comprehension. Some generative models\cite{lee2022muse, zhao2019multi, westny2024diffusion} including GANs and VAEs have also been applied to trajectory prediction tasks to better capture the multi-modal nature of future trajectories. Multi-Path\cite{chai2019multipath} uses a set of dense trajectory anchors to capture the agent intentions. CoverNet\cite{phan2020covernet} formulates trajectory prediction as classifying among a set of dense trajectories. These generative methods require a very large number of samples to cover all possible trajectories, especially the ones with low probability that are not necessarily unimportant.

\begin{table*}[tbp]
  \centering
  \begin{tabular}{c c c c c c}
    \toprule[1.2pt]
    \makebox{\textbf{\thead{Method}}} & \makebox{\textbf{\thead{Interaction(tar-sur)}}} & \makebox{\textbf{\thead{Interaction(sur-sur)}}} & \makebox{\textbf{\thead{Integration capability}}} & \makebox{\textbf{\thead{Considers dynamics}}} &
    \makebox{\textbf{\thead{Ego agent's motion}}} 
    \\
    \midrule[1.2pt]
    S-LSTM (CVPR) \cite{alahi2016social} & \checkmark & & & & \\
    C-LSTM (CVPR) \cite{deo2018convolutional} & \checkmark & & & & \\
    PiP (ECCV) \cite{song2020pip} & \checkmark & & \checkmark & & \checkmark \\
    CF-LSTM (ICRA) \cite{xie2021congestion} & \checkmark & & & & \\
    WSiP (AAAI) \cite{wang2023wsip} & \checkmark & \checkmark & & & \\
    BAT (AAAI) \cite{liao2024bat} & \checkmark & \checkmark & & & \\
    C2F-TP (AAAI) \cite{wang2024c2f} & \checkmark & \checkmark & & \checkmark & \\
    \midrule[0.5pt]
    \textbf{Ours} & \checkmark & \checkmark & \checkmark & \checkmark & \checkmark \\
    \bottomrule[1.2pt]
  \end{tabular}
  \caption{A summary of recent state-of-the-art trajectory forecasting methods. \textbf{Interaction(tar-sur)} and \textbf{Interaction(sur-sur)} respectively denote that the proposed method takes into account the interactions between the target agent and surrounding agents, as well as the interactions within the group of surrounding agents. \textbf{Integration capability} denotes the model can be easily integrated into the robotic system. \textbf{Considers dynamics} implies that the method factors in the intention changes during the movement of agents, thereby dynamically modeling social interaction. \textbf{Ego agent's motion} signifies that the method accounts for the potential motion of the ego agent.}
  \label{table1}
\end{table*}

\subsection{Interaction-Aware Trajectory Prediction}
The movements of an agent is not only influenced by its past trajectory but also related to other neighboring agents. Numerous works\cite{song2020pip, chen2022intention, wang2023wsip, chen2022intention} have been proposed on how to model the social interaction between agents. A classic study proposed early is the social force model (SFM)\cite{helbing1995social} which reflects the interaction between the agents by attractive and repulsive forces. Social LSTM\cite{alahi2016social} extracts the historical trajectory features of each agent with LSTMs separately and discovers the interactions between neighboring agents using social pooling structure. Convolution Social Pooling\cite{deo2018convolutional} improves upon this by incorporating convolution layers, which better captures the spatial connections between different agents. INT2\cite{yan2023int2} presents a large-scale interactive trajectory dataset for interactive trajectory prediction. SocialCircle\cite{wong2024socialcircle} builds a new angle-based social interaction representation. In this work, we propose a novel cross time domain intention-interaction model for trajectory prediction. Compared to previous methods, our method fuses information across different time domains and spatial scales, empowering the network with stronger feature representation capabilities.

\subsection{Conditional Trajectory Prediction}
Conditional prediction methods\cite{huang2022tip, sun2022m2i, ngiam2021scene, liu2021deep} explore the relationship between candidate trajectories of the ego agent and future trajectories of other surrounding agents. Trajectron++\cite{salzmann2020trajectron++} incorporates heterogeneous data and future motions in predictions. Precog\cite{rhinehart2019precog} and PiP\cite{song2020pip} attempts to incorporate the future trajectory of the ego agent to explore how surrounding agents would be affected when the ego agent take different candidate trajectories. M2I\cite{sun2022m2i} focuses on agent pairs, where one of the agents is predicted as influencer and the other as reactor. Then it utilizes a conditional model to generate consistent future trajectories for the reactor based on the influencer's marginal future predictions. Different from existing conditional prediction methods, our proposed model only requires rough instead of precise future trajectory, which can be easily obtained even in non-autonomous system scenarios. For instance, a rough left turn trajectory can be inferred based on a vehicle turning on left turn signals, and a roughly straight trajectory can be inferred for a vehicle driving in a straight lane.

\begin{figure*}[tbp]
  \centering
   \includegraphics[width=1.0 \linewidth]{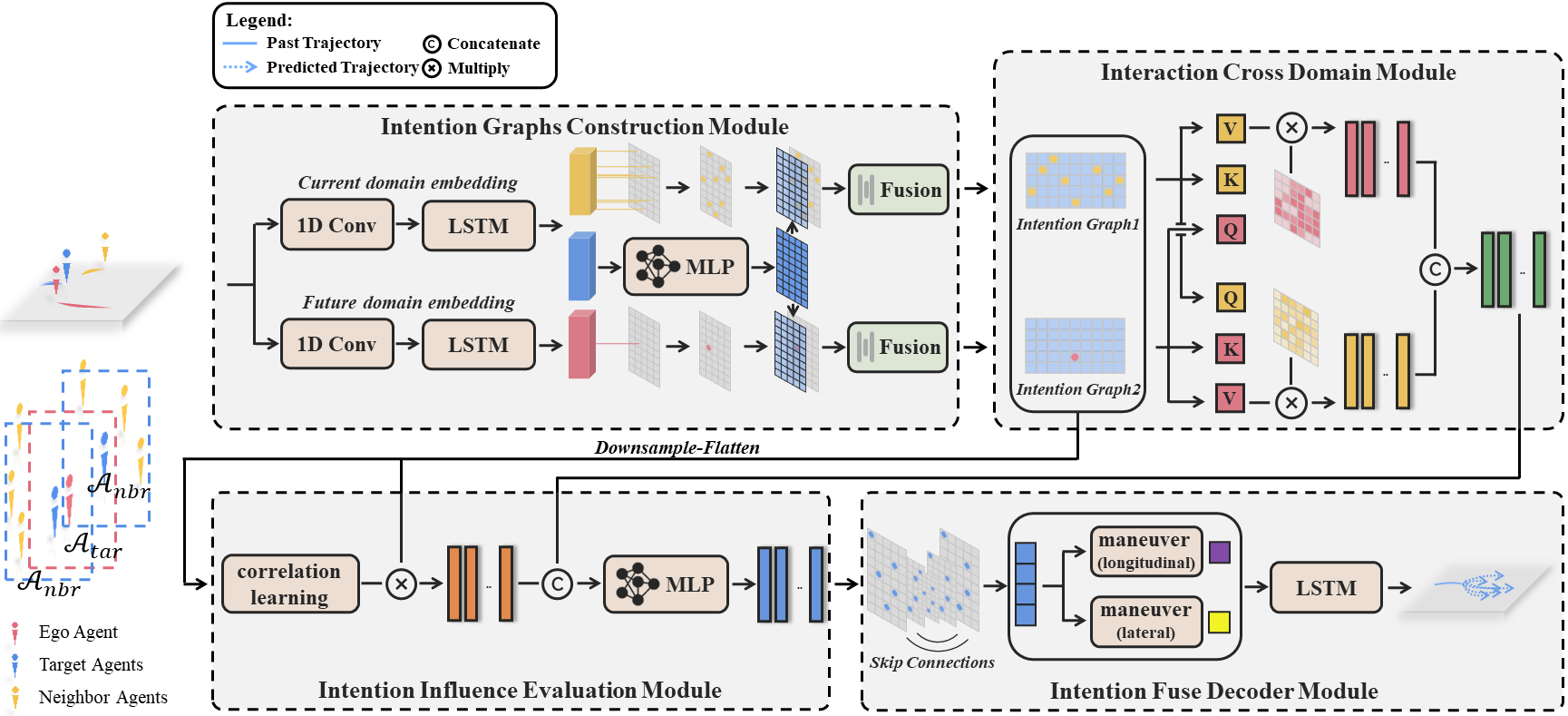}
   \caption{The overview of our proposed method: For each predicted target, two intention graphs are first constructed. Cross-domain interactions between these graphs are then modeled to refine behavior intention. Subsequent prioritized evaluation of other agents' influences further enhances intention estimation. Finally, multi-scale feature fusion and decoding generate diverse plausible trajectories.}
   \label{Fig.2}
\end{figure*}

\section{Problem Formulation}
Considering human-robot interactive autonomous system scenarios, the ego agent is commanded by planning and control modules, while the perception module detects surrounding agents around the ego agent. We formulate the multi-agent trajectory prediction problem as predicting future trajectories of all surrounding target agents given historical trajectories of surrounding agents within a certain range and future motion of the ego agent. The objective is to learn the posterior distribution $P\left(\mathcal{Y}|\mathcal{\mathcal{X},\mathcal{L}}\right)$ of multiple targets' future trajectories $\mathcal{Y}=\{{Y}_{i}|{v}_{i}\in {V}_{tar}\}$, where ${V}_{tar}$ is the set of predicted targets selected within an ego-vehicle-centered area ${\mathcal{A}}_{tar}$. The $\mathcal{X}=\{{X}_{i}|{v}_{i}\in {V}_{tar}\}$ denotes the past trajectories of the target agent. At any time $t$, the history trajectory and future trajectory of an agent $i$ are denoted as ${X}_{i}=\{{x}_{i}^{t-{T}_{obs}+1}, {x}_{i}^{t-{T}_{obs}+2}..., {x}_{i}^{t}\}$ and ${Y}_{i}=\{{y}_{i}^{t+1}, {y}_{i}^{t+2}..., {y}_{i}^{t+{T}_{pred}}\}$, where the elements of ${x}_{i}$,${y}_{i}$$\in {\mathbb{R}}^{2}$ represent coordinates in the past and future, respectively, while ${T}_{obs}$ and ${T}_{pred}$ refer to the number of frames for observation and prediction. We also consider the setting where we condition on the ego-agent's future motion plan, for example when evaluating responses to a set of motion primitives. In this setting, we additionally assume that we know the ego-agent's future motion plan for the next $T$ timesteps, $\mathcal{L}={Y}_{ego}=\{{y}_{ego}^{t+1}, {y}_{ego}^{t+2}..., {y}_{ego}^{t+{T}_{pred}}\}$.

\section{Motivation}
In this work we are interested in developing a trajectory prediction method that (1) directly designed for real-world robotic systems and can be integrated with downstream modules such as decision making and motion planning; (2) reasons the target agent's intention changes over time to more accurately infer its future trajectory choices; and (3) models social interaction in greater detail, selectively distinguishing the varying degrees of influence that different agents have on the target agent's future trajectory.

By introducing the potential motion of the ego agent, this method can evaluate the future trajectories of surrounding target agents under different potential motions of the ego agent, thereby providing a basis for downstream modules of the robotic system. It is worth re-emphasizing that we merely require a coarse potential motion of the target agent, rather than precise and complete planned trajectory. Additionally, we find that incorporating the potential motion of the ego agent enables the modeling of social interactions between the ego agent and target agents in the future time domain. This enables us to jointly analyze intention information across both the current and future time domains, achieving fusion of feature information in spatial dimensions, temporal dimensions, and among different agents, thereby obtaining more accurate future intentions for the target agent. Finally, the model also learns the varying degrees of influence that different agents have on the target's future trajectory selection, enabling a hierarchical consideration of the interactions between the target agent and its surroundings, and achieving a more granular modeling process.

\section{Proposed method}

Our proposed method CiT is summarized in Figure \ref{Fig.2}, which processes each predicted agent in its own centric area ${\mathcal{A}}_{nbr}$. Thus, the area ${\mathcal{A}}_{nbr}$ contains three types of agents in total: the ego agent, the target agent, and the other neighboring agents. CiT contains four modules in total. Sec.\ref{4.2} proposes an Intention Graphs Construction Module to construct behavior intentions of the target agent in two time domains, Sec.\ref{4.3} proposes an Interaction Cross Domain Module to dynamically model social interactions across time domains, Sec.\ref{4.4} proposes an Intention Influence Evaluation Module to assess which agent's influence the target agent should focus more on, and Sec.\ref{4.5} proposes an Intention Fusion Decoder Module to fuse features and generate multiple possible future trajectories for each target agent.

\subsection{Intention Graphs Construction}
\label{4.2}
First, the trajectories ${X}_{ego}$, ${X}_{tar}$, ${X}_{nbr}$ are encoded by the 1D-CNN followed by the long short term memory (LSTM) model. Since the trajectories of target agent, neighbor agents and ego agent belong to different time domains, CNN-LSTMs with different weights are adopted. While 1D-CNN-LSTM can extract temporal properties, it can hardly capture the spatial interactions between target agent and its surrounding agents. Therefore, we build a grid centered at the location of the target agent and adopt the convolutional social pooling layer proposed by \cite{deo2018convolutional}. The ego agent encoding ${\xi}_{ego}$ is placed to a social tensor in the target-centric grid with respect to its locations, then convolution and pooling layers are applied on the social tensor to capture social context for the spatial interaction between target agent and ego agent in the future time domain. A similar processing method is also adopted for the neighbor agents encoding ${\xi}_{nbr}$ to capture the social context for the spatial interaction between the target agent and neighbor agents in the current time domain. The target encoding ${\xi}_{tar}$ is concatenated with the two time domain social contexts to generate two Intention Graphs ${\mathcal{V}}^{c},{\mathcal{V}}^{f}\in{\mathbb{R}}^{H\times W\times (D+1)}$, where $H$ and $W$ are the height and width of the target-centric grid ${\mathcal{A}}_{nbr}$ respectively, and $D$ is the dimension after trajectory encoding. 
\begin{equation}
{\xi}_{tar}=LSTM (1D-CNN({X}_{tar}))    
\end{equation}
\begin{equation}
{\xi}_{ego}=LSTM (1D-CNN({X}_{ego}))    
\end{equation}
\begin{equation}
{\xi}_{nbr}=LSTM (1D-CNN({X}_{nbr}))
\end{equation}
\subsection{Interaction Cross Domain}
\label{4.3}
Both Intention Graphs already contain the historical trajectory information of the target agent and social interaction information with surrounding agents. Therefore, each of the intention graph has a certain capability of estimating the future trajectory of the target agent. However, taking the humans' theory of mind that they can navigate through various scenarios without collisions with surroundings as a reference, the future trajectory inferred unilaterally from intention in one time domain is inadequate. Because humans often comprehensively consider social behaviors in different time domains before selecting trajectories, which requires the model to jointly analyze behavior intentions in different time domains. Starting from the completion of Intention Graphs Construction Module, ${\mathcal{V}}^{c}$ is transformed to generate the current time domain intention matrix ${\mathcal{M}}_{c}=[{\mathcal{V}}_{1}^{c},{\mathcal{V}}_{2}^{c},...,{\mathcal{V}}_{D+1}^{c}]$, with ${V}_{d}^{c}\in {\mathbb{R}}^{HW}$ and similarly for the future time domain intention matrix ${\mathcal{M}}_{f}$ is generated from ${\mathcal{V}}^{f}$, where ${\mathcal{M}}_{c},{\mathcal{M}}_{f}\in {\mathbb{R}}^{HW\times(D+1)}$. The Intention Graph in the current time domain proposes Queries to the Intention Graph in the future time domain, and explores which interactions in the future time domain will affect its current Behavior Intention through the Key and Value of the future time domain Intention Graph, so as to revise its own Intention. The Intention Graph in the future time domain will also obtain an Intention that is more consistent with real human thoughts by proposing Queries to the Intention Graph in the current time domain. To conduct cross time domain intention analysis jointly, we propose the formulation of the cross time domain representation with:

\begin{equation}
{\widetilde{\mathcal{M}}}_{c}={\mathcal{F}}^{trans}({\mathcal{M}}_{c},{\mathcal{M}}_{f},{\mathcal{M}}_{f})
\end{equation}

\begin{equation}
{\widetilde{\mathcal{M}}}_{f}={\mathcal{F}}^{trans}({\mathcal{M}}_{f},{\mathcal{M}}_{c},{\mathcal{M}}_{c})
\end{equation}

${\widetilde{\mathcal{M}}}_{c},{\widetilde{\mathcal{M}}}_{f}\in {\mathbb{R}}^{HW\times {D}^{\prime}}$, and the ${\mathcal{F}}^{trans}(Q,K,V)$ is defined as:

\begin{equation}
{\mathcal{F}}^{trans}(Q,K,V)={\mathcal{F}}^{fc}({\mathcal{F}}^{attn}(Q,K,V))
\end{equation}




Through the Interaction Cross Domain Module, the Intention Graphs in the two time domains are fully integrated with each other's information and adjust their own intentions accordingly. This allows the network to have the capability of dynamically modeling social interactions across time domains. We concatenate the ${\widetilde{M}}_{c}$ and ${\widetilde{M}}_{f}$ to obtain $\mathcal{I}$, a more precise representation of the target agent's intention which explores the correlation between social interaction in the two time domains.

\subsection{Intention Influence Evaluation}
\label{4.4}

The Intention Influence Evaluation Module learns the influence degree of social interactions in the two time domains on the target agent, which is a more delicate modeling of social interaction behaviors, allowing the network to consider social interactions in a prioritized manner. We flatten the downsampled Intention Graphs ${\mathcal{V}}^{c}$ and ${\mathcal{V}}^{f}$ and concatenate them with the encoding of the target agent to generate social contexts in the two time domains, which are then fed into fully connected and softmax layers to produce two weights $\beta_{1} $ and $\beta_{2} $ representing the importance of social information in each time domain respectively. The two weights are multiplied with the social contexts in the two time domains respectively to obtain vector $\mathcal{G}$. Finally, $\mathcal{G}$ is concatenated with the vector $\mathcal{I}$ obtained from the Interaction Cross Domain Module to generate the intention representation $\mathcal{Z}$ of the target agent.

Note that we multiply the weights of different time domains with the low-level Intention Graphs, rather than the deeper and more abstract Intention Graphs generated after Interaction Cross Domain. We want to make use of the shallow information contained in the low-level Intention Graphs through this method. At the same time, we want the model to remember to some extent the "original" intention generated from social interactions in its own time domain, preventing the model from over-focusing on social behaviors in the other time domain thus causing unnecessary intention correction. Before feeding it into the decoder, the Intention $\mathcal{Z}$ at this point already contains the social interaction information of the target agent with different agents across time domains and spaces. In addition, it incorporates both high-level and low-level intention features.

\subsection{Intention Fusion Decoder}
\label{4.5}
In order to predict future trajectories for all target agents around the ego agent, each target agent intention represented as $\mathcal{Z}$ is placed into an ego agent-centric grid based on its location, resulting in a social intention tensor $\mathcal{S}$. To further extract interaction features at different spatial scales, we apply a fulley convolutional network (FCN) structure on the social intention tensor $\mathcal{S}$ and obtain the intention feature ${\mathcal{Z}}^{+}$. We refer to the method of \cite{deo2018convolutional} and predefine six maneuver categories $\mathcal{C}=\{{c}_{k}|k=1,2,...,6\}$ in advance to address the inherent multi-modality of human behavior by predicting the distribution for each of the maneuver classes along with the probability for each maneuver class. The predefined maneuver categories including three lateral maneuvers (lane keeping, left lane change, right lane change) and two longitudinal maneuvers (normal driving and braking). We feed each target agent's intention feature ${\mathcal{Z}}^{+}$ into fully connected layers followed by softmax layers that output the lateral and longitudinal maneuver probabilities. These can be multiplied to give the value of each maneuver $P({c}_{k}|\mathcal{L},X)$. The lateral and longitudinal maneuver class are transformed into one-hot vectors and concatenated with the intention feature ${\mathcal{Z}}^{+}$. Then the resulted feature vector is fed into LSTM layers to generate the predicted location, which can be represented by a bivariate Gaussian distribution over ${T}_{pred}$ frames:

\begin{equation}
{\hat{y}}_{i}^{t+{T}_{pred}}\thicksim \mathcal{N}({\mu }_{i}^{t+{T}_{pred}},{\sigma }_{i}^{t+{T}_{pred}},{\rho }_{i}^{t+{T}_{pred}})
\end{equation}

where the mean vector is the sum of all displacements along the future frames ${T}_{pred}$ with the location at the last frame $t$, the standard deviation vector ${\sigma }_{i}^{t+{T}_{pred}}\in {\mathbb{R}}^{2}$ and the correlation coefficient ${\rho }_{i}^{t+{T}_{pred}}\in \mathbb{R}$. Finally, the posterior probability of all surrounding target agents' future trajectories could be estimated from:

\begin{equation}
P(\mathcal{Y}|\mathcal{X},\mathcal{L})=\!\!\!\!\displaystyle\prod_{{v}_{i}\in {V}_{tar}}^{}\displaystyle\sum_{k=1}^{\left | \mathcal{C}\right |}{P}_{{\theta }_{i}}({Y}_{i}|{c}_{k},\mathcal{X},\mathcal{L})P({c}_{k}|\mathcal{X},\mathcal{L})
\end{equation}

where the Gaussian parameters for all future frames of target agent ${v}_{i}$ is written as ${\theta }_{i}$.

\section{Experiments}
Following existing methods\cite{wang2023wsip, song2020pip, deo2018convolutional, alahi2016social, zhao2019multi, gupta2018social, gao2023dual, wang2020multi}, we evaluate our proposed method on two public-available datasets NGSIM and HighD. For each dataset, 70\% is split for training and 10\%, 20\% for validation and test. The objective is to predict the future 5s (25 frames) behavioral trajectories for all target agents surrounding the ego agent using the 3s (15 frames) past trajectory and the ego agent's potential motion. In addition, to demonstrate our proposed method only requires the rough potential motion of the ego agent, ego agent's future trajectory is downsampled to 1Hz (5 frames) in the testing and evaluation.

\subsection{Implementation Details}
Each data instance contains an agent specified as the ego agent. The target agents whose trajectories we want to predict are all agents located in the ego-agent-centric area ${\mathcal{A}}_{tar}$. The area ${\mathcal{A}}_{tar}$ has a size of 200$\times $35 feet, discretized into a 25$\times $5 spatial grid. For each target agent, the operating area ${\mathcal{A}}_{nbr}$ has the same size as ${\mathcal{A}}_{tar}$. Ideally, we want to minimize the negative log-likelihood over all training instances.

\begin{equation}
-log(\displaystyle\sum_{{v}_{i}\in {V}_{tar}}^{}{P}_{\theta }({Y}_{i}|{c}_{k},\mathcal{X},\mathcal{L})P({c}_{k}|\mathcal{X},\mathcal{L}))
\end{equation}

However, each training instance only corresponds to the one true maneuver class that was actually performed. Thus we minimize the negative log likelihood of the predictive distribution associated with the true maneuver.

\begin{equation}
-\displaystyle\sum_{{v}_{i}\in {V}_{tar}}^{}log({P}_{\theta }({Y}_{i}|{c}_{true},\mathcal{X},\mathcal{L})P({c}_{true}|\mathcal{X},\mathcal{L}))
\end{equation}

For the incorporation of the ego agent's potential motion $\mathcal{L}$, we use the ego agent's actual future trajectory over the prediction horizon 5s (25 frames) as input during training. In evaluation and testing, we downsample the ego agent's actual trajectory to 5 frames to verify our model only requires the ego agent's "rough intention". The maintained validity of our model when only "rough intention" is available illustrates the high robustness and convenience of real-world deployment of our proposed method. Meanwhile, the introduction of the ego agent's potential motion in our model can effectively combine with downstream trajectory planning and control tasks, having high application value.

All the experiments were conducted on NVIDIA RTX 4090(24GB). We employed 1D-conv to upsample the input of the model from dimension 2 to 32. The encoder LSTM has 64 dimensional state while the decoder has a 128 dimensional state. In these experiments, all parameter settings are aligned with the compared methods. We use the leaky-ReLU activation with $\alpha =0.1$ for all layers.

\begin{table*}[tbp]
  \centering
  \scalebox{1.138}{
  \begin{tabular}{c|c c c c c c c c c c|c}
    \toprule[1.2pt]
    \makebox{\thead{Time}} & \makebox{\thead{S-LSTM \\ \cite{alahi2016social}}} & \makebox{\thead{C-LSTM \\ \cite{deo2018convolutional}}} & \makebox{\thead{MATF-GAN \\ \cite{zhao2019multi}}} & \makebox{\thead{SAMMP \\ \cite{mercat2020multi}}} & \makebox{\thead{PiP \\ \cite{song2020pip}}} & \makebox{\thead{CF-LSTM \\ \cite{xie2021congestion}}} & \makebox{\thead{Flash \\ \cite{antonello2022flash}}} & \makebox{\thead{WSiP \\ \cite{wang2023wsip}}} & \makebox{\thead{BAT \\ \cite{liao2024bat}}} & \makebox{\thead{C2F-TP \\ \cite{wang2024c2f}}} & \thead{Ours} \\
    \midrule[1.2pt]
    1s & 0.59/2.10 & 0.58/1.96 & 0.66 & \underline{0.51} & 0.55/\underline{1.72} & 0.55 & \underline{0.51} & 0.56/1.77 & \textbf{0.23} & \underline{0.32} & 0.43/\textbf{1.61} \\
    
    2s & 1.29/3.66 & 1.27/3.46 & 1.34 & 1.13 & 1.18/\underline{3.30} & \underline{1.10} & 1.15 & 1.23/\underline{3.30} & \textbf{0.81} & 0.92 & \underline{0.88}/\textbf{3.14} \\
    
    3s & 2.13/4.61 & 2.11/4.32 & 2.08 & 1.88 & 1.94/\underline{4.17} & \underline{1.78} & 1.84 & 2.05/4.17 & \underline{1.54} & 1.62 & \textbf{1.47}/\textbf{3.95} \\
    
    4s & 3.21/5.37 & 3.19/4.95 & 2.97 & 2.81 & 2.88/\underline{4.80} & 2.73 & \underline{2.64} & 3.08/\underline{4.80} & 2.52 & \underline{2.44} & \textbf{2.36}/\textbf{4.54} \\
    
    5s & 4.55/5.99 & 4.53/5.48 & 4.13 & 3.67 & 4.04/\underline{5.32} & 3.82 & \underline{3.62} & 4.34/\underline{5.32} & 3.62 & \underline{3.45} & \textbf{3.23}/\textbf{5.04} \\
    
    avg & 2.35/4.35 & 2.34/4.03 & 2.24 & 2.00 & 2.12/\underline{3.86} & 2.00 & \underline{1.95} & 2.25/3.87 & 1.74 & 1.75 & \textbf{1.67}/\textbf{3.66} \\
    \bottomrule[1.2pt]
  \end{tabular}
  }
  \caption{Comparison with baseline models on NGSIM dataset. RMSE/NLL are reported. Lower is better. \textbf{Bold}/\underline{underlined} fonts represent the best/second-best result. Our method achieves the best performance in RMSE/NLL.}
  \label{table2}
\end{table*}

\begin{table*}[tbp]
  \centering
  \scalebox{1.138}{
  \begin{tabular}{c|c c c c c c c c c c|c}
    \toprule[1.2pt]
    \makebox{\thead{Time}} & \makebox{\thead{S-LSTM \\ \cite{alahi2016social}}} & \makebox{\thead{C-LSTM \\ \cite{deo2018convolutional}}} & \makebox{\thead{S-GAN \\ \cite{gupta2018social}}} & \makebox{\thead{N-LSTM \\ \cite{messaoud2019non}}} & {\thead{PiP \\ \cite{song2020pip}}} & \makebox{\thead{M-LSTM \\ \cite{messaoud2020attention}}} & \makebox{\thead{DLM \\ \cite{khakzar2020dual}}} & \makebox{\thead{DRBP \\ \cite{gao2023dual}}} & \makebox{\thead{WSiP \\ \cite{wang2023wsip}}} & \makebox{\thead{C2F-TP \\ \cite{wang2024c2f}}} & \thead{Ours} \\
    \midrule[1.2pt]
    1s & 0.21/0.46 & 0.24/0.43 & 0.30 & 0.20 & 0.17/\textbf{0.14} & 0.19 & 0.22 & 0.41 & 0.20/0.31 & \textbf{0.11} & \underline{0.12}/\underline{0.15} \\
    
    2s & 0.65/2.55 & 0.68/2.54 & 0.78 & 0.57 & 0.52/\underline{2.24} & 0.55 & 0.61 & 0.79 & 0.60/2.31 & \textbf{0.41} & \textbf{0.41}/\textbf{2.14} \\
    
    3s & 1.31/3.81 & 1.26/3.72 & 1.46 & 1.14 & 1.05/\underline{3.48} & 1.10 & 1.16 & 1.11 & 1.21/3.51 & \underline{0.92} & \textbf{0.85}/\textbf{3.31} \\
    
    4s & 2.16/4.67 & 2.15/4.51 & 2.34 & 1.90 & 1.76/\underline{4.33} & 1.84 & 1.80 & \textbf{1.40} & 2.07/4.32 & 1.64 & \underline{1.49}/\textbf{4.11} \\
    
    5s & 3.29/5.35 & 3.31/5.13 & 3.41 & 2.91 & 2.63/\underline{4.99} & 2.78 & 2.80 & - & 3.14/4.95 & \underline{2.60} & \textbf{2.43}/\textbf{4.74} \\
    
    avg & 1.52/3.37 & 1.53/3.27 & 1.66 & 1.34 & 1.23/\underline{3.04} & 1.29 & 1.32 & - & 1.44/3.08 & \underline{1.14} & \textbf{1.06}/\textbf{2.89} \\
    \bottomrule[1.2pt]
  \end{tabular}
  }
  \caption{Comparison with baseline models on HighD dataset. RMSE/NLL are reported. Lower is better. \textbf{Bold}/\underline{underlined} fonts represent the best/second-best result. Our method achieves the best or second-best performance in RMSE/NLL.}
  \label{table3}
\end{table*}

\begin{table*}[tbp]
  \centering
  \scalebox{1.08} {
  \begin{tabular}{c | c c c c c | c c c c c | c}
    \toprule[1.2pt]
    \makebox{Method} & \makebox{info (c)} & \makebox{info (f)} & \makebox{ICD} & \makebox{IIE} & \makebox{Fusion} & \makebox{1s} & \makebox{2s} & \makebox{3s} & \makebox{4s} & \makebox{5s} & \makebox{avg} \\
    \midrule[1.2pt]
    Variant1 &  &  &  &  &  & 0.68/2.14 & 1.66/3.81 & 2.96/4.76 & 4.56/5.42 & 5.44/6.03 & 3.06/4.43 \\
    
    Variant2 & \checkmark &  &  &  & \checkmark & 0.51/\textbf{1.60} & 1.09/3.19 & 1.79/\underline{3.98} & 2.69/4.62 & 3.83/5.15 & 1.98/3.71 \\
    
    Variant3 &  & \checkmark &  &  & \checkmark & 0.54/1.70 & 1.32/3.27 & 2.31/4.21 & 3.51/4.85 & 4.90/5.35 & 2.52/3.88 \\
    
    Variant4 & \checkmark & \checkmark &  &  & \checkmark & 0.53/1.83 & 1.13/3.34 & 1.81/4.17 & 2.64/4.76 & 3.66/5.25 & 1.95/3.87 \\
     
    Variant5 & \checkmark & \checkmark & \checkmark &  & \checkmark & \underline{0.46}/1.64 & \textbf{0.88}/\underline{3.16} & \underline{1.56}/\underline{3.98} & \underline{2.50}/\underline{4.56} & \underline{3.41}/\underline{5.06} & \underline{1.76}/\underline{3.68} \\
    
    Ours & \checkmark & \checkmark & \checkmark & \checkmark & \checkmark & \textbf{0.43}/\underline{1.61} & \textbf{0.88}/\textbf{3.14} & \textbf{1.47}/\textbf{3.95} & \textbf{2.36}/\textbf{4.54} & \textbf{3.23}/\textbf{5.04} & \textbf{1.67}/\textbf{3.66} \\
    \bottomrule[1.2pt]
  \end{tabular}
  }
  \caption{Ablation study conducted on NGSIM dataset. info(c) refers to the trajectory information of the surrounding agents in the current time domain, info(f) refers to the potential motion of the ego agent in the future time domain. ICD stands for Interaction Cross Domain Module, IIE stands for Intention Influence Evaluation, and Fusion means performing feature fusion through the Fully Connected Network (FCN) structure.}
  \label{table4}
\end{table*}

\begin{table*}[tbp]
  \centering
  \scalebox{1.08}{
  \begin{tabular}{c | c c c c c | c c c c c | c}
    \toprule[1.2pt]
    \makebox{Method} & \makebox{info (c)} & \makebox{info (f)} & \makebox{ICD} & \makebox{IIE} & \makebox{Fusion} & \makebox{1s} & \makebox{2s} & \makebox{3s} & \makebox{4s} & \makebox{5s} & \makebox{avg} \\
    \midrule[1.2pt]
    Variant1 &  &  &  &  &  & 0.22/0.55 & 0.65/2.65 & 1.32/3.94 & 2.22/4.87 & 3.43/5.59 & 1.57/3.52 \\
    
    Variant2 & \checkmark &  &  &  & \checkmark & 0.20/0.28 & 0.62/\underline{2.27} & 1.30/3.47 & 2.23/4.31 & 3.42/4.96 & 1.55/3.06 \\
    
    Variant3 &  & \checkmark &  &  & \checkmark & 0.21/0.34 & 0.65/2.34 & 1.33/3.52 & 2.23/4.33 & 3.35/4.95 & 1.55/3.10 \\
    
    Variant4 & \checkmark & \checkmark &  &  & \checkmark & 0.22/0.20 & 0.61/2.31 & 1.19/3.44 & 1.96/4.28 & 2.95/4.88 & 1.39/3.02 \\
     
    Variant5 & \checkmark & \checkmark & \checkmark &  & \checkmark & \underline{0.17}/\textbf{0.14} & \underline{0.46}/\underline{2.27} & \underline{0.96}/\underline{3.42} & \underline{1.50}/\underline{4.24} & \underline{2.55}/\underline{4.86} & \underline{1.13}/\underline{2.99} \\
    
    Ours & \checkmark & \checkmark & \checkmark & \checkmark & \checkmark & \textbf{0.12}/\underline{0.15} & \textbf{0.41}/\textbf{2.14} & \textbf{0.85}/\textbf{3.31} & \textbf{1.49}/\textbf{4.11} & \textbf{2.43}/\textbf{4.74} & \textbf{1.06}/\textbf{2.89} \\
    \bottomrule[1.2pt]
  \end{tabular}
  }
  \caption{Ablation study conducted on HighD dataset. info(c) refers to the trajectory information of the surrounding agents in the current time domain, info(f) refers to the potential motion of the ego agent in the future time domain. ICD stands for Interaction Cross Domain Module, IIE stands for Intention Influence Evaluation, and Fusion means performing feature fusion through the Fully Connected Network (FCN) structure.}
  \label{table5}
\end{table*}

\subsection{Comparison with SOTA Methods}
We measure the performance of different trajectory prediction methods using two metrics: RMSE and NLL. 1) RMSE calculates the deviation between predicted trajectories and ground truth future trajectories. For multiple predicted trajectories, we evaluate RMSE using the trajectory under the maneuver with highest probability. 2) The negative log-likelihood (NLL) of true trajectories under the predictive distribution fitted by the model is also adopted in evaluation, because RMSE tends to average all prediction results and has limitations for multi-modal trajectory prediction.

\noindent\textbf{NGSIM dataset.} We compare our method with the state-of-the-art prediction methods at different timestamps; see Table \ref{table2}. The method we proposed has achieved either the first or the second place in the metrics at all timestamps, and it outperforms the current state-of-the-art (SOTA) methods in terms of the average metrics of RMSR and NLL. We have reduced the RMSE and NLL from 1.74/3.86 to 1.67/3.66. We have achieved improvements of \textbf{4.02\%} and \textbf{5.18\%} in the RMSE and NLL metrics respectively.

\noindent\textbf{HighD dataset.} We compare our method with the state-of-the-art prediction methods at different timestamps; see Table \ref{table3}. The method we proposed has achieved either the first or the second place in the metrics at all timestamps, and it outperforms the current state-of-the-art (SOTA) methods in terms of the average metrics of RMSR and NLL. We have reduced the RMSE and NLL from 1.14/3.04 to 1.06/2.89. We have achieved improvements of \textbf{7.02\%} and \textbf{4.93\%} in the RMSE and NLL metrics respectively.

\begin{figure*}[tbp]
  \centering
   \includegraphics[scale=0.3]{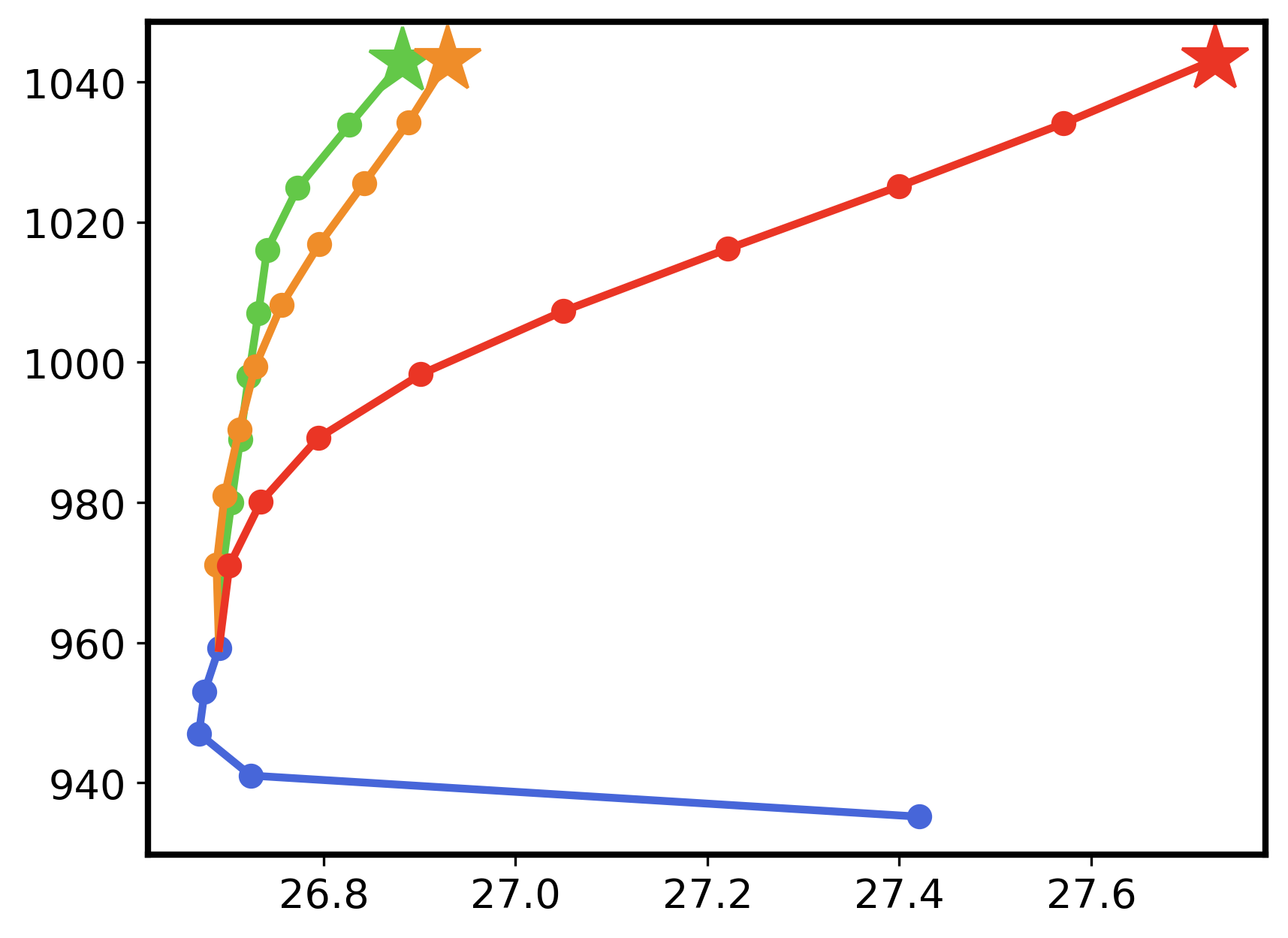}
   \includegraphics[scale=0.3]{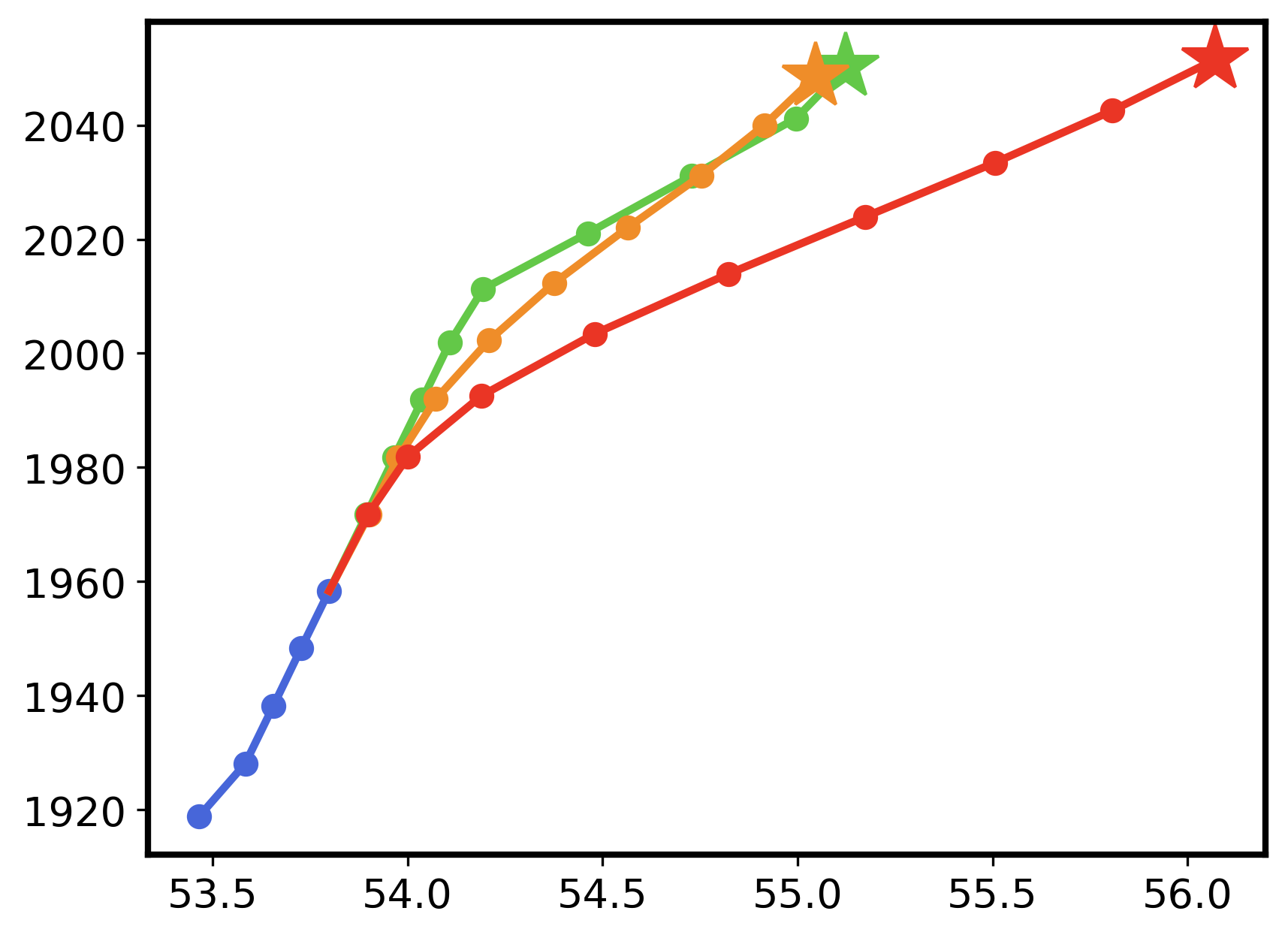}
   \includegraphics[scale=0.3]{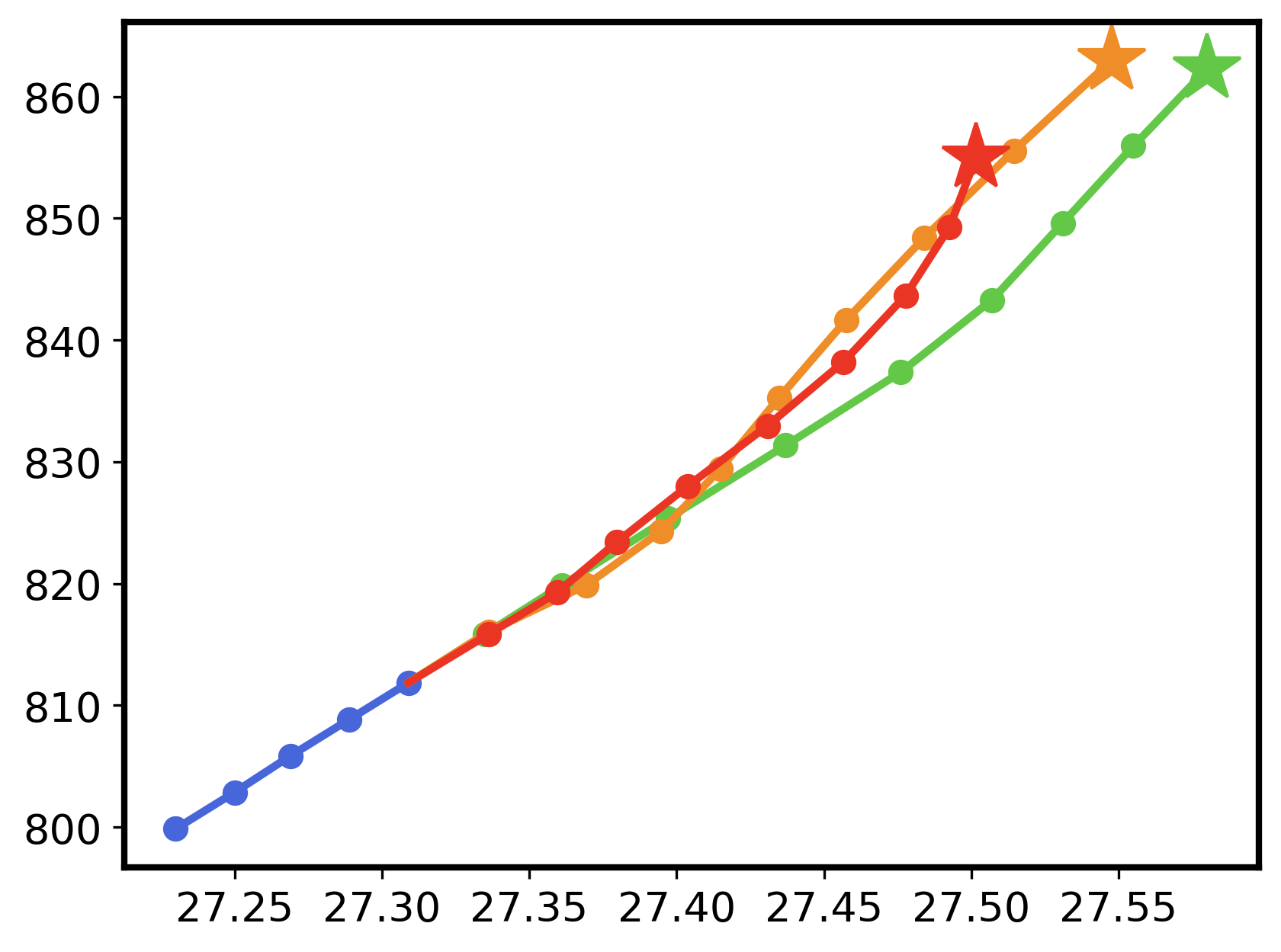}
   \includegraphics[scale=0.3]{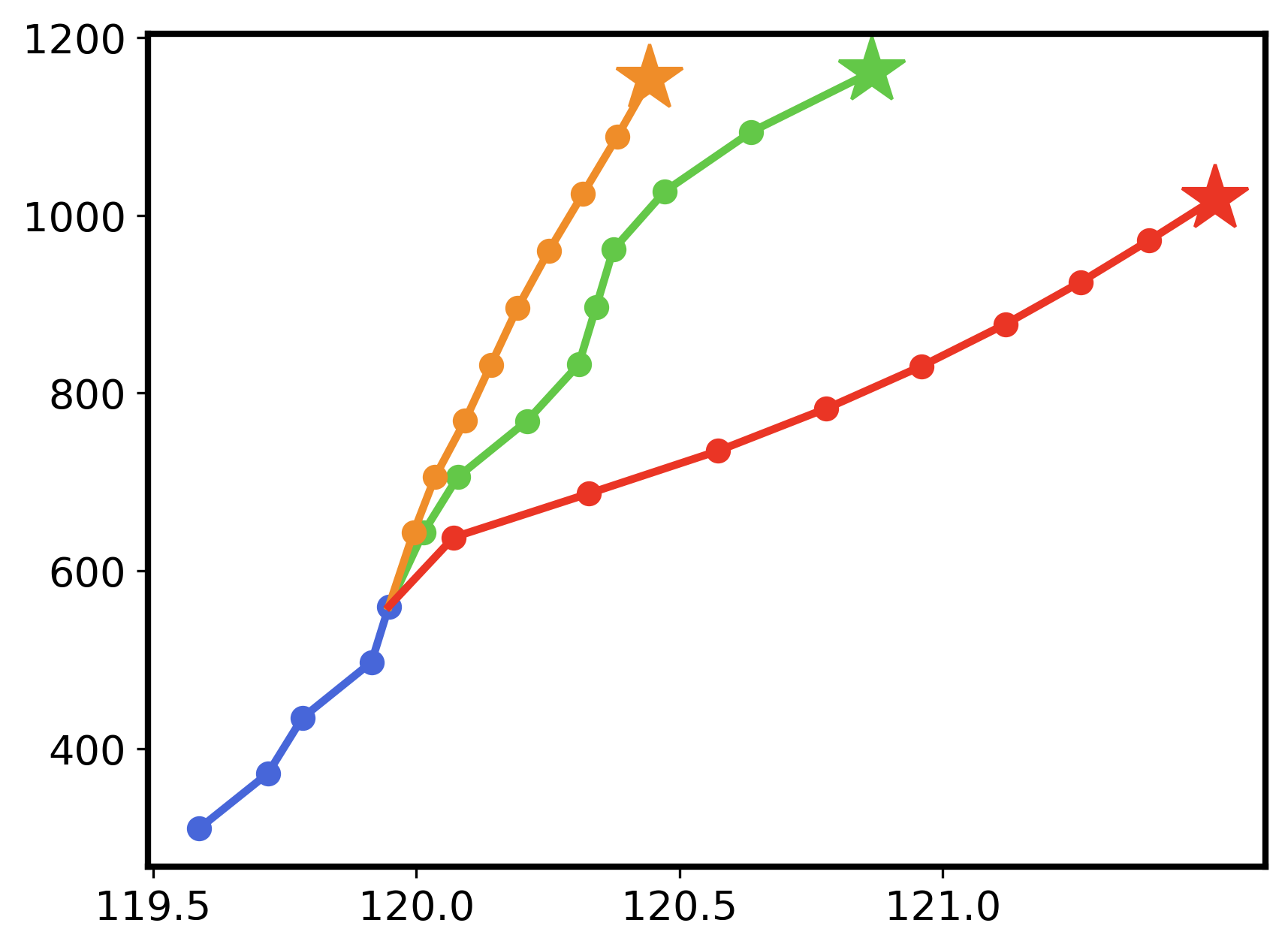}
   \includegraphics[scale=0.3]{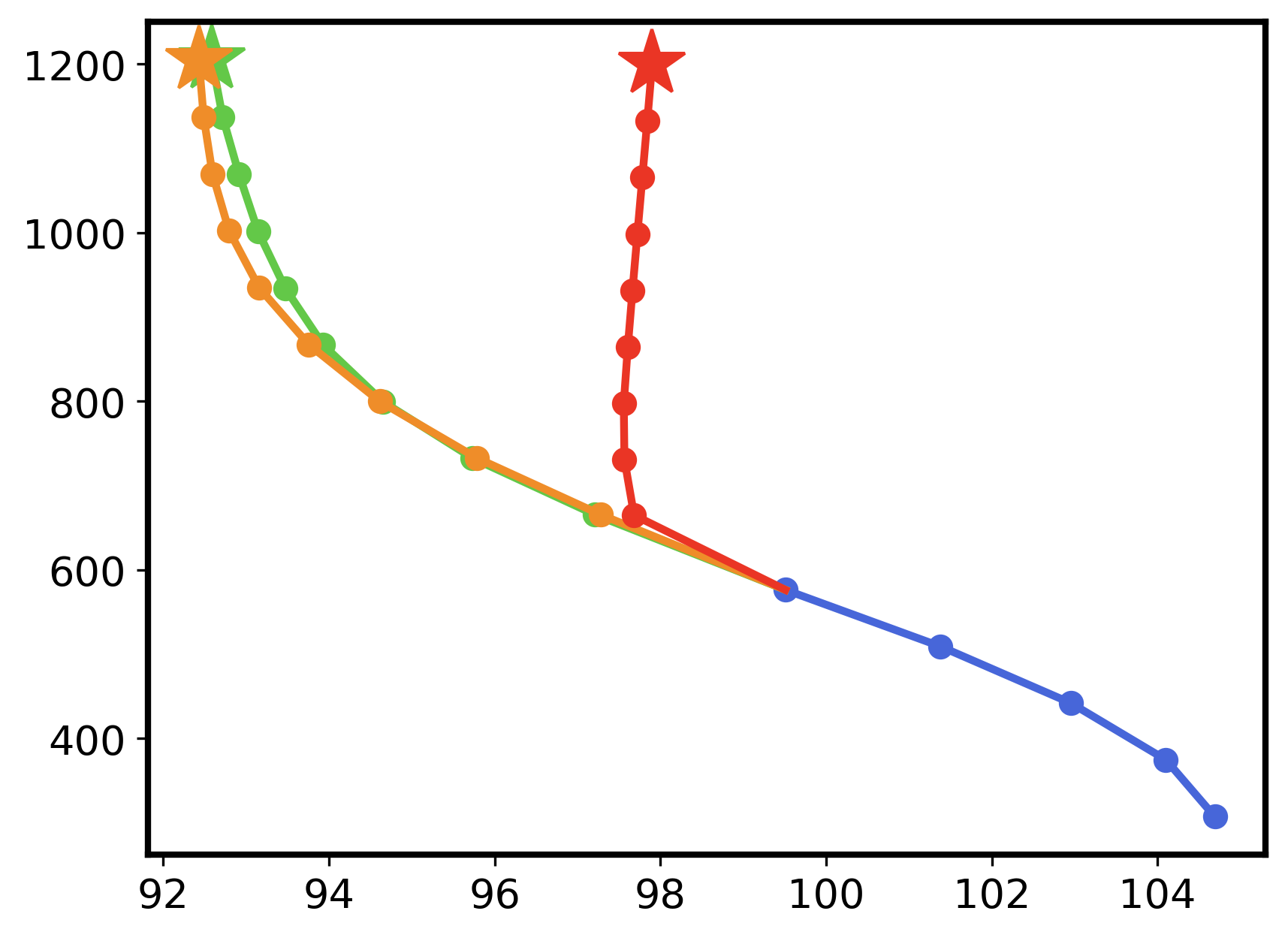}
   \includegraphics[scale=0.3]{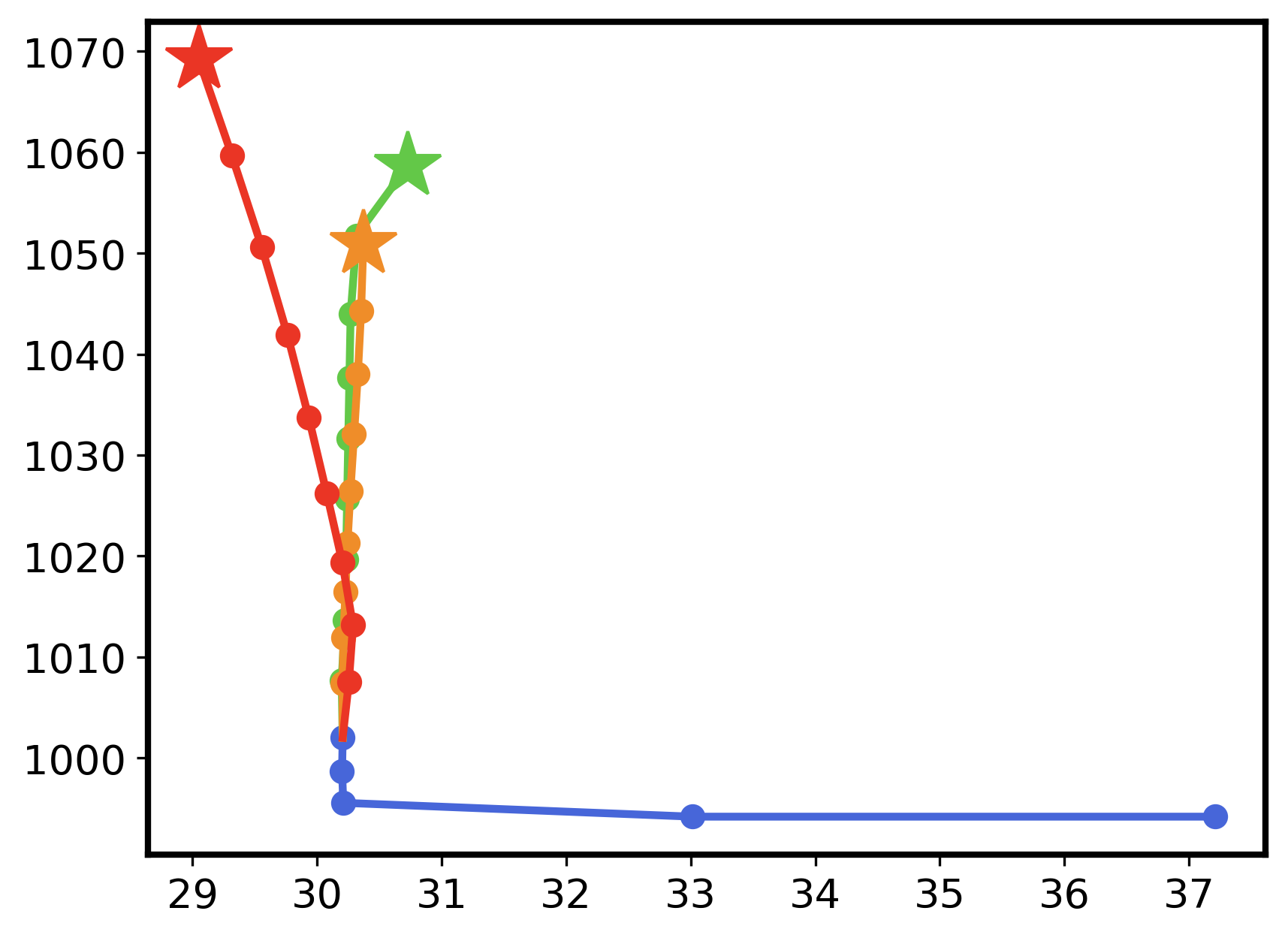}
   \includegraphics[scale=0.3]{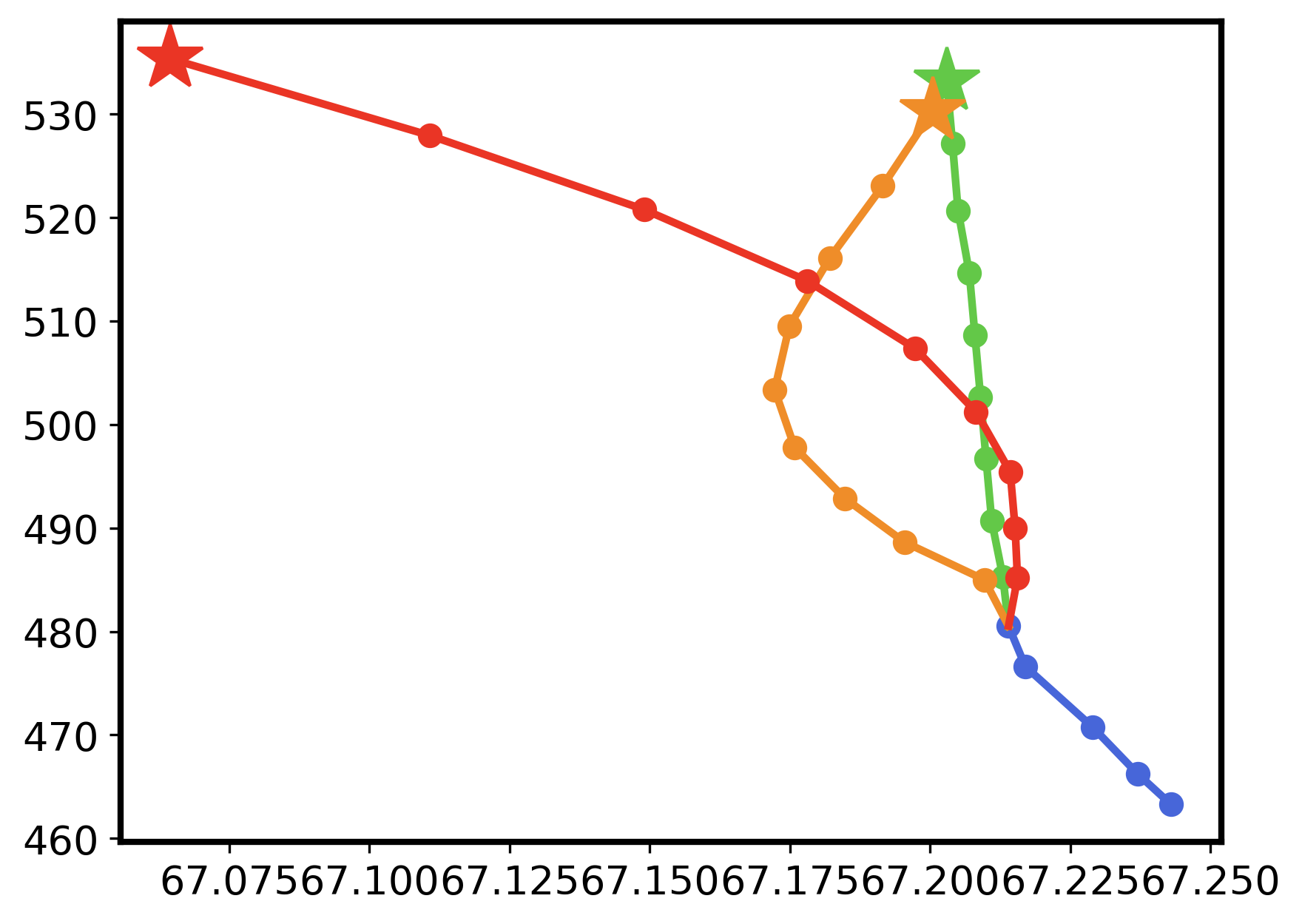}
   \includegraphics[scale=0.3]{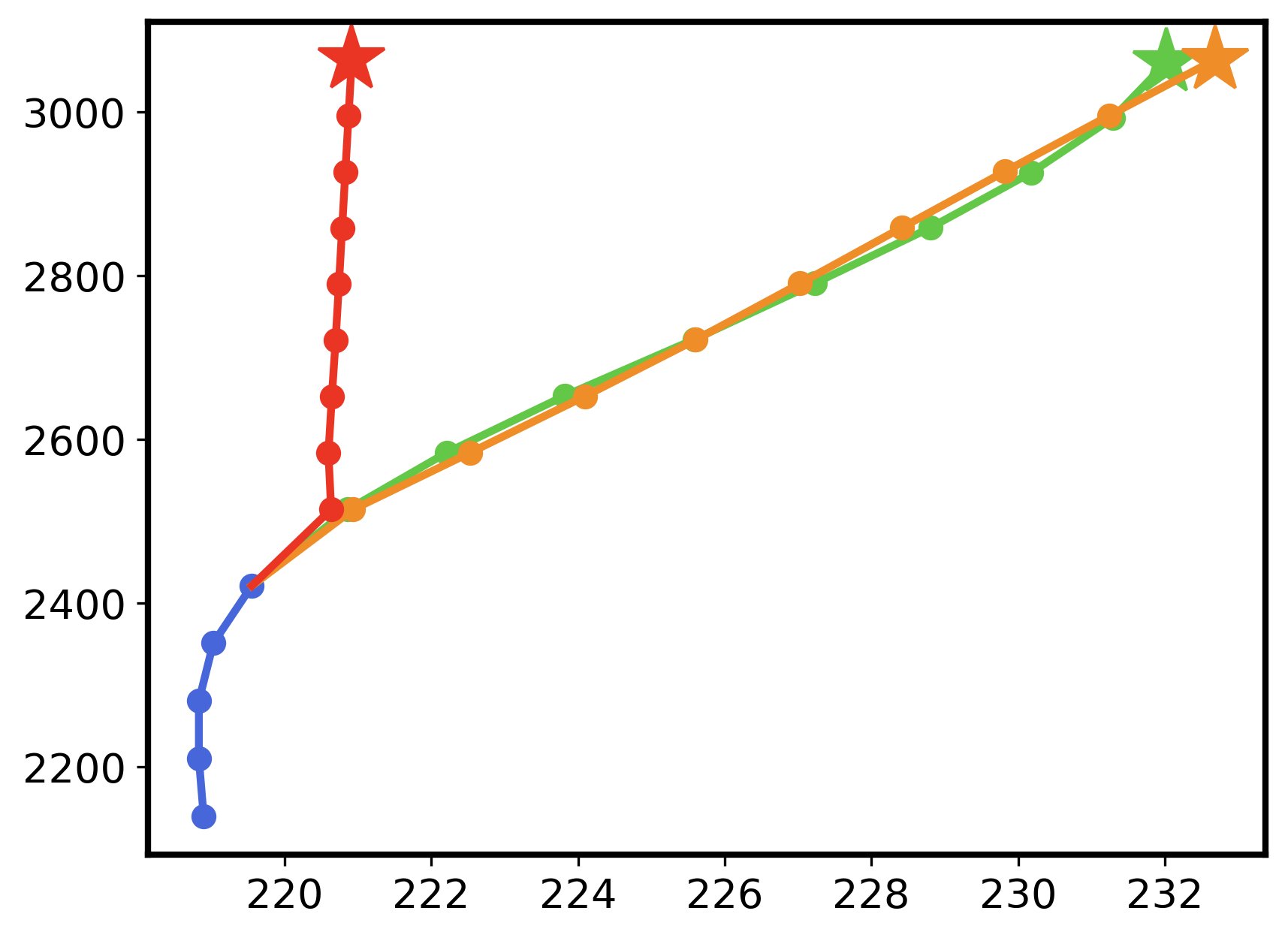}
   \caption{Visualization comparison on NGSIM and HighD. Given the past trajectories(blue), we illustrate the ground truth trajectories(green) and predicted trajectories by CiT(yellow) and PiP(red). Best viewed in color and zoom-in for more clarity.}
   \label{Fig.3}
\end{figure*}

\begin{figure*}[tbp]
  \centering
   \includegraphics[scale=0.3]{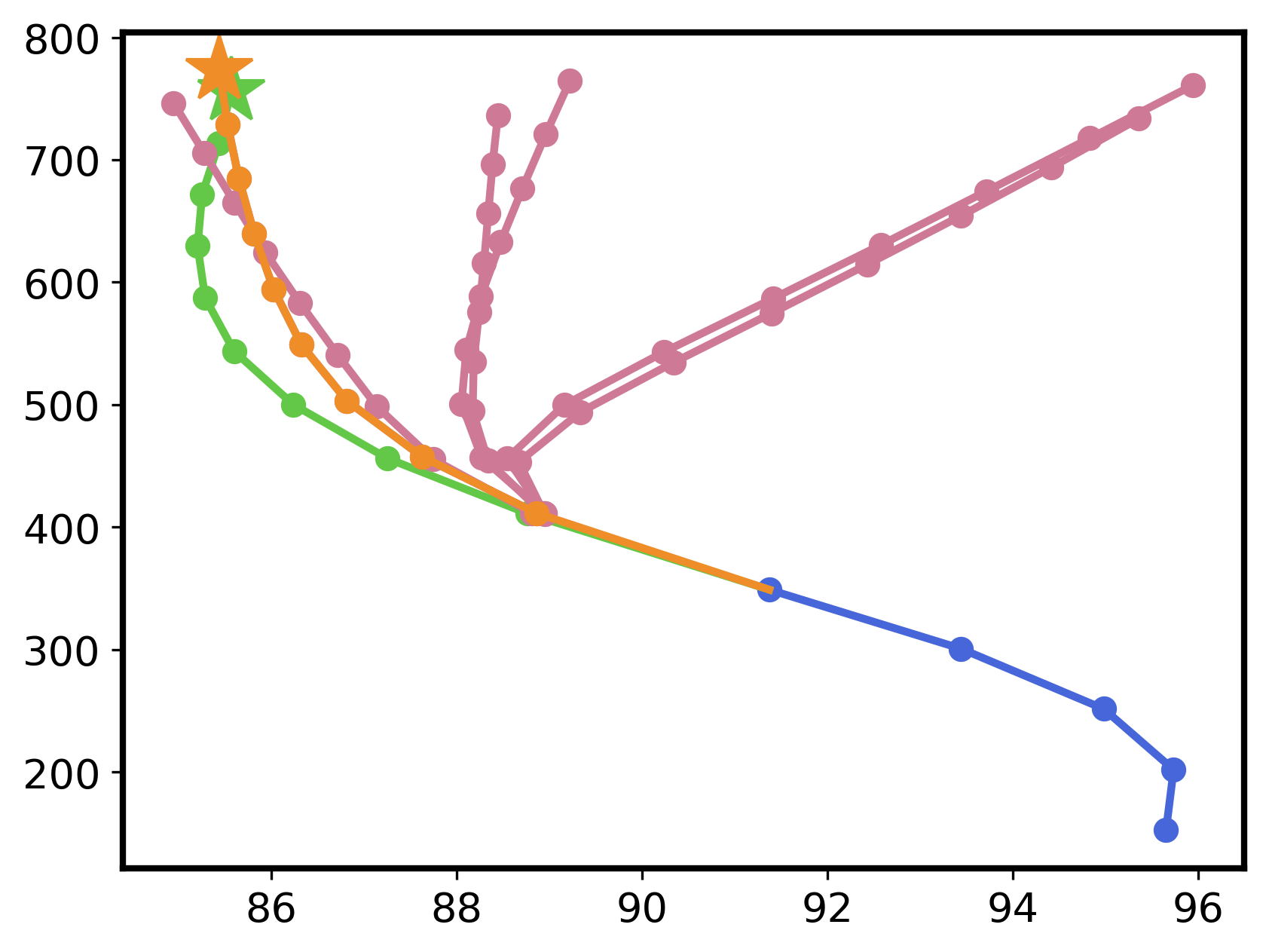}
   \includegraphics[scale=0.3]{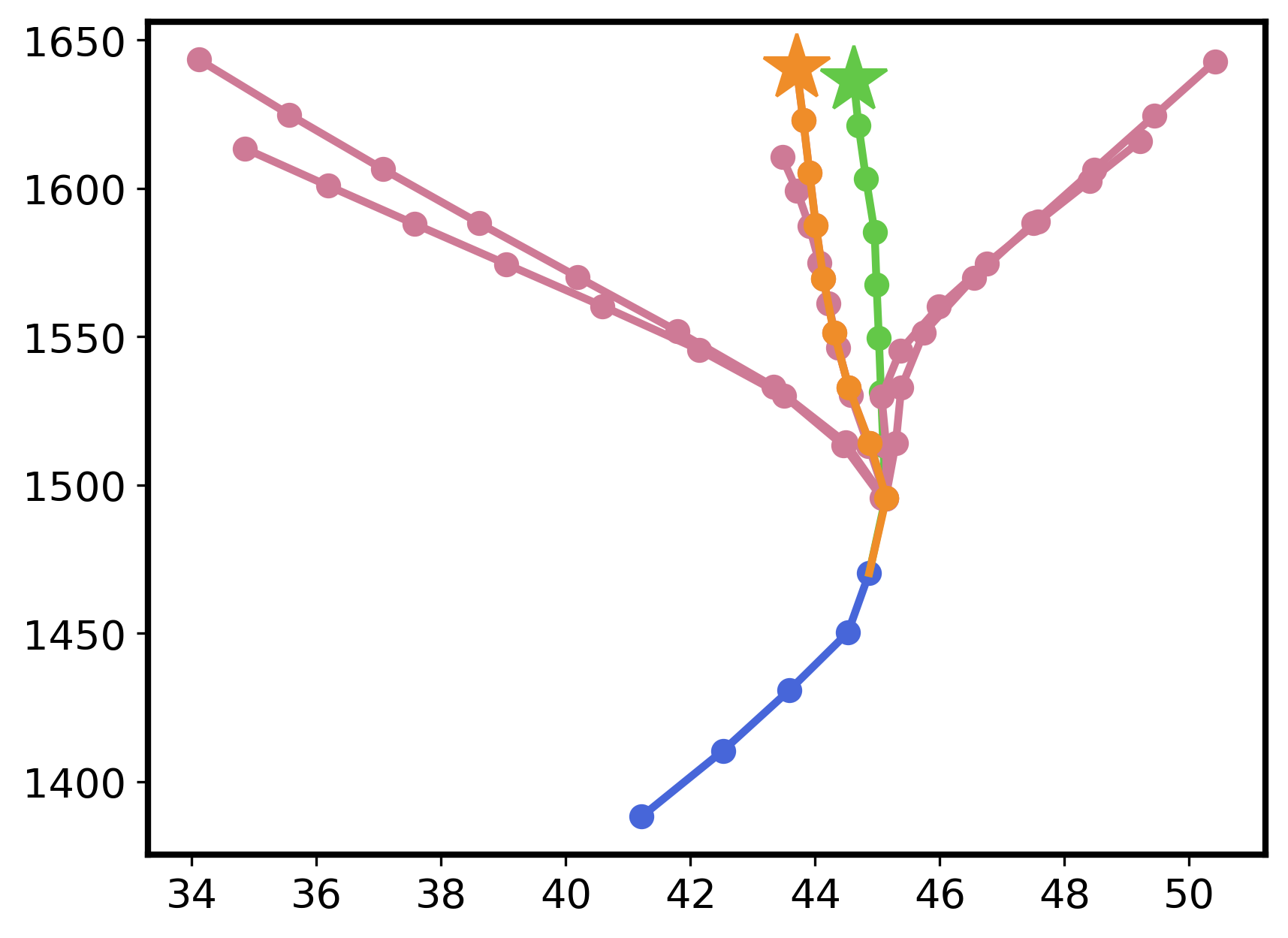}
   \includegraphics[scale=0.3]{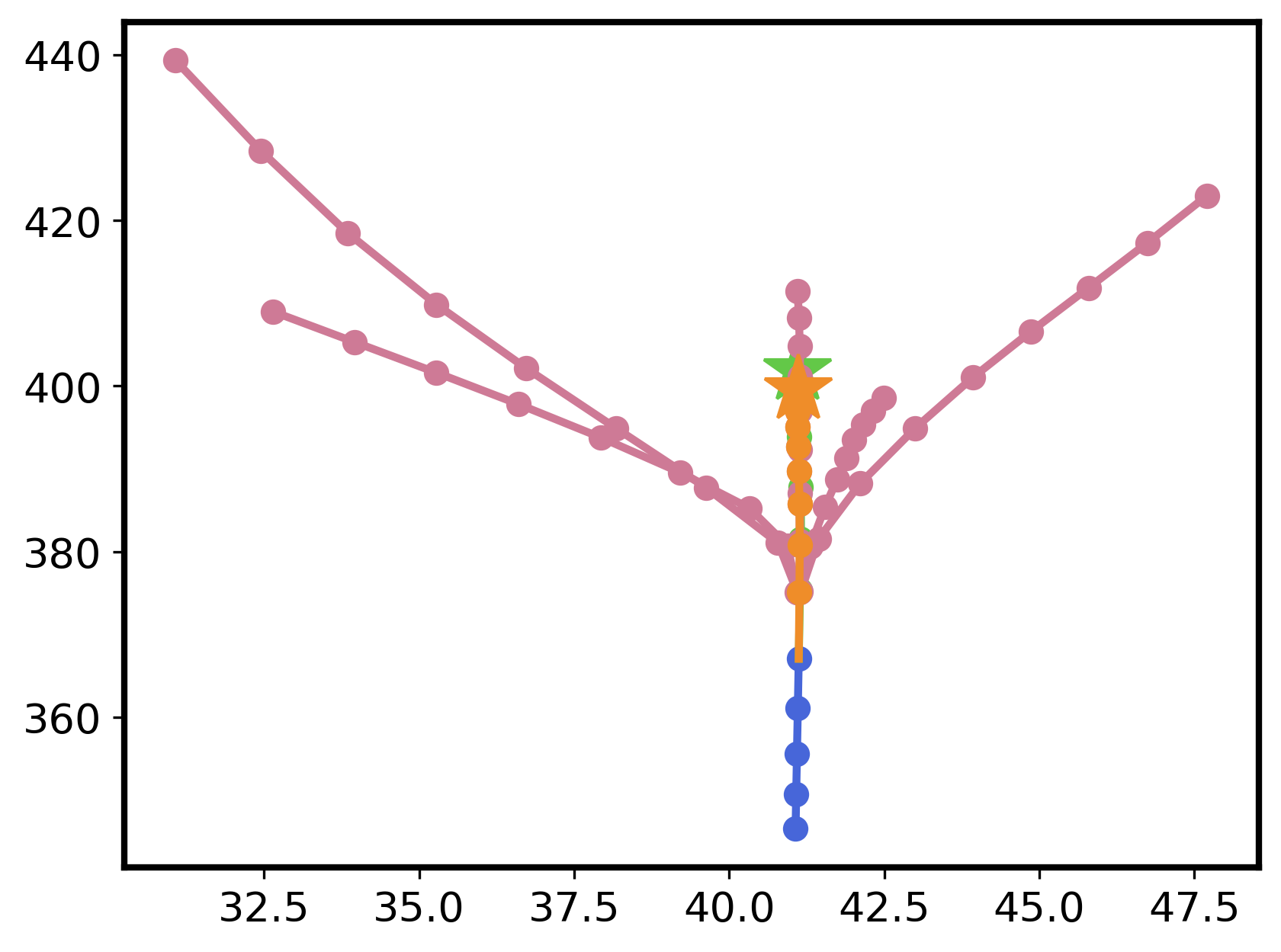}
   \includegraphics[scale=0.3]{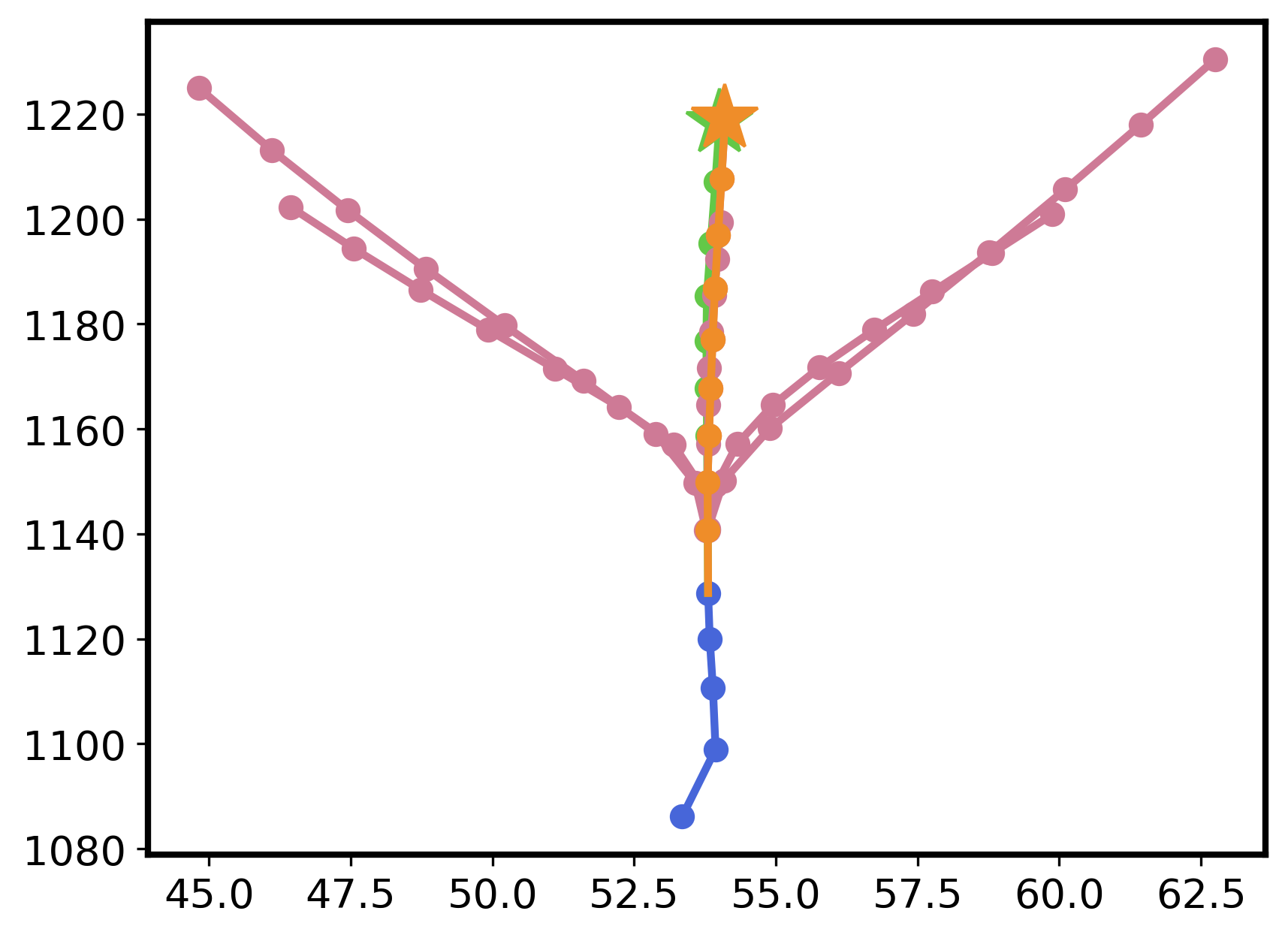}

   \caption{Visualization multi-modal prediction on NGSIM. Given the past trajectory(blue), we illustrate the ground truth trajectory(green), the predicted prediction(yellow), which corresponds to the maneuver with highest probability among all predicted trajectories, and the other predictions(purple). Best viewed in color and zoom-in for more clarity.}
   \label{Fig.4}
\end{figure*}
\subsection{Qualitative Results}

\textbf{Visualization of predicted trajectory.} Figure \ref{Fig.3} compares the predicted trajectories of the previous conditional trajectory prediction state-of-the-art method, PiP(red), our CiT(yellow), historical trajectories(blue) and the ground-truth(green). In the first row, it is shown that although the previous state-of-the-art methods can correctly capture the motivation of the target agent, that is, the trend of the motion direction is correct, our prediction results are closer to the ground truth. In the second row, we can observe that the previous methods fail to correctly capture the motivation of the target agent. For example, in the first column, the historical trajectory of the target agent is changing lanes to the left and will continue to move leftward. However, PiP predicts that it will move straight after changing lanes to the left. In practical application scenarios, such misjudgment of the motion motivation is extremely dangerous. From this, we can see that by mutually correcting the intentions across the time domain, not only can more accurate trajectory information be obtained, but it is even possible to correct the originally mispredicted motivations.

\noindent\textbf{Visualization of multi-modal prediction.} Figure \ref{Fig.4} shows the multiple future trajectories predicted by our proposed method based on predefined maneuvers. The figure \ref{Fig.4} contains multiple predicted future trajectories under the six pre-defined motivations in Sec.\ref{4.5}. The yellow trajectory represents the future trajectory predicted under the motivation with the highest probability by the method, which is the prediction result output by the method. The purple trajectories are the future trajectories predicted under other motivations with relatively lower probabilities. We can see that the method can capture agent maneuvers like lane changing(first column), braking(third column) and normal driving(fourth column). In some challenging scenarios, for example, in the first column, the historical trajectory of the agent shows a lane change to the left, and the agent will continue to move leftward for a certain distance before driving straight. It is relatively difficult for the method to predict the straight trajectory after the lane change to the left. In the second column, the entire historical trajectory of the agent is changing lanes to the right. However, precisely within the prediction time window, the agent has completed the lane change and maintains a straight driving motion. In such a situation, the model often predicts that the agent will continue to change lanes to the right based on the information of the rightward lane change in the historical trajectory. Nevertheless, as can be seen from the figure \ref{Fig.4}, the method we proposed can correctly capture the information that the agent has completed the lane change and will maintain a straight driving motion.

\subsection{Ablation Studies}
Tables \ref{table4} and \ref{table5} show the contribution of each component added to model over the baseline. Variant1 is the baseline, which only uses LSTM to predict future trajectories based on past trajectories. Variant2 takes surrounding agents' state as conditional input, and uses fully convolutional networks(FCN). Variant3 takes ego agent's potential motion as conditional input and also utilizes the FCN structure. To demonstrate that the performance improvement achieved by the ICD module stems from the mutual complementarity of semantic intentions across time domains rather than simply increasing the parameters, we introduce Variant4, which employs self-attention mechanism for information fusion and shares almost the same number of parameters as the ICD module. Variant5 is based on Variant4, replacing the self-attention method to integrate information with the Interaction Cross Domain Module(ICD). Our method adds the Intention Influence Evaluation Module(IIE) on Variant5. The experimental results show that performance improves as we add the proposed modules. Compared to Variant2 and Variant3 which model social interactions unilaterally from either the current or future time domain, Variant4 achieves better results by integrating information from the two time domains. Moreover, Variant5 further improves over Variant4 by using social information from the other time domain to correct the intention in its own time domain, and conducts joint analysis of social interactions across time domains. It can be observed that prediction accuracy is improved with the IIE module. This indicates the model can differentiate the influence across time domains to refine social interaction modeling, and incorporating more global, shallow intention information can further enhance model performance.

\section{Conclusion}
This paper presents a novel trajectory prediction method that models cross-time-domain social interactions to capture temporal intention dynamics. Our approach incorporates four key components: potential motion integration, intention graph construction, cross-domain interaction modeling, and intention influence assessment. These designs enable inter-temporal information exchange and intention refinement, while the integration of ego-agent potential motion facilitates seamless integration with downstream planning and control modules in robotic systems.

\newpage
\section*{Acknowledgements}
The study is supported by the National Key R\&D Program of China (No. 2023YFB4301900).
\bibliographystyle{ACM-Reference-Format}
\balance
\bibliography{main}

\end{document}